\documentclass[letterpaper, 10 pt, conference]{ieeeconf}
\IEEEoverridecommandlockouts
\usepackage{amsmath,amsfonts,amssymb}
\usepackage{bm}
\usepackage{algorithmic}
\usepackage{algorithm}
\usepackage{array}
\usepackage{graphicx}
\usepackage{adjustbox}
\usepackage{mathtools}
\usepackage{cite}
\usepackage{booktabs}
\usepackage{multirow}
\usepackage{threeparttable}
\usepackage[table]{xcolor}
\usepackage{url}
\usepackage{stfloats}
\usepackage[font=small,skip=3pt,labelfont=bf]{caption}
\usepackage{tikz}
\usetikzlibrary{arrows.meta,positioning,fit,backgrounds,calc}
\usepackage[colorlinks=true, linkcolor=blue, citecolor=blue, urlcolor=blue]{hyperref}

\tikzset{
  fl/.style  ={-{Latex[length=1.6mm,width=1.2mm]}, line width=0.6pt,
               draw=arrowsteel},
  blkbase/.style={draw, rounded corners=1.4pt, align=center,
                  font=\sffamily\scriptsize, inner xsep=2.4pt,
                  inner ysep=2.6pt, line width=0.45pt},
  io/.style  ={blkbase, fill=black!5,        draw=black!45},
  s1/.style  ={blkbase, fill=blue!8,         draw=blue!55!black},
  s2/.style  ={blkbase, fill=orange!14,      draw=orange!65!black},
  sel/.style ={blkbase, fill=red!7,          draw=red!55!black},
  res/.style ={blkbase, fill=green!9,        draw=green!45!black},
  lbl/.style ={font=\sffamily\tiny, text=black!60},
  sw/.style  ={blkbase, fill=black!4, draw=black!45, text width=22mm},
}

\definecolor{arrowsteel}{RGB}{88,124,168}  
\definecolor{tabhead}{gray}{0.93}
\definecolor{tabbest}{RGB}{158,190,224}
\definecolor{tabsecond}{RGB}{214,228,242}

\newif\ifrtmark
\rtmarkfalse

\newcommand{\meanerr}{\bar{e}}
\newcommand{\Mtwo}{\mathcal{M}_2}
\newcommand{\Mthree}{\mathcal{M}_3}
\newcommand{\medres}{\tilde{e}}          
\newcommand{\sighat}{\hat{\sigma}}       
\newcommand{\eval}{\meanerr_{\mathrm{v}}}
\newcommand{\emaxv}{e^{\max}_{\mathrm{v}}}

\title{\LARGE \bf
Automatic Reproducible Camera Intrinsic Calibration}

\newif\ifanon
\anonfalse
\ifanon
  \author{Anonymous Authors%
  \thanks{Affiliation withheld for double-anonymous review.}%
  }
\else
  \author{Xiangcheng Hu%
  \thanks{X. Hu is with the Department of Electronic and Computer Engineering,
  The Hong Kong University of Science and Technology, Hong Kong SAR, China
  {\tt\small (xhubd@connect.ust.hk)}}%
  }
\fi

\begin{document}
\maketitle
\thispagestyle{empty}
\pagestyle{empty}

\begin{abstract}
Accurate camera intrinsic calibration is fundamental to robot perception,
and the accuracy depends on the quality of the
collected images. However, existing target-based calibration methods often
require the practitioner to manually filter out high-quality images and to
specify an appropriate radial distortion order. 
This paper presents a fully
automatic intrinsic calibration pipeline that determines both from the
collected data. We adopt an iterative rejection scheme that estimates
parameters on a candidate image set and removes views whose mean residual
exceeds a multiple of the median. Crucially, this process runs independently
under each candidate distortion order, so that the retained image set is
consistent with the residual scale of that order. 
Further, the distortion order is selected on held-out images, with the intrinsics and distortion fixed and only the board pose re-estimated, ensuring that an added coefficient is supported by independent observations.
Finally, we integrate both steps into an interactive calibration
tool that supports full-pipeline data inspection and parameter estimation.
Experiments on our own camera data and five public real-world datasets show that image filtering reduces the held-out reprojection error by 25\%, the order
selection further by 5\%, achieving the lowest held-out mean among four
compared configurations without manual image selection. We will release the
code and data to facilitate future research.
\end{abstract}


\section{Introduction}
\label{sec:intro}

\subsection{Motivation and Challenges}
\label{sec:challenges}

Cameras are the primary sensor of autonomous driving and embodied
intelligence systems, and all geometric computation derived from the images,
including depth estimation, pose estimation, visual odometry, and multi-sensor fusion~\cite{hu2024paloc}, relies on accurate intrinsics~\cite{schops2020generic}. The intrinsics are calibrated once from a small set of planar target images and then held
fixed throughout the deployment, so a calibration error recurs in every
downstream application as a systematic geometric bias~\cite{hu2025mapeval}. Target-based
calibration has been solved in essentially the same form for more than two
decades, and the solver itself is
mature~\cite{zhang2000flexible,brown1971close,matlab_calibrator,ros_camera_calibration,mrcal}.
However, the bottleneck is no longer the solver but the two inputs that
precede it: which images are selected for estimation, and what order of radial
distortion model is adopted. Both are typically set by convention or operator
experience. Corner extraction under motion blur, large incidence angles and
low resolution carries view-dependent systematic
components~\cite{tang2017precision}, and since each view carries only six
board-pose parameters, such error propagates into the intrinsics shared by
the whole set. A higher-order distortion model
necessarily reduces the calibration-set residual, yet this reduction is an
inherent property of fitting more parameters, not evidence of better
prediction on unseen images. While production calibration controls data
quality at collection through fixtures and prescribed
orientations, field calibration with a hand-held board must assess quality
from the collected data itself, where the only available quantity is the
reprojection error, which cannot resolve either problem:
\begin{itemize}
  \setlength{\itemsep}{1pt}\setlength{\parskip}{0pt}\setlength{\leftmargini}{1.1em}
  \item \textbf{Which images to estimate from.} Per-view residuals are
        available only after a parameter has been estimated, and the views
        that should be removed have already displaced that calibration. A
        filtering criterion independent of the fit is therefore required
        first, after which the residual scale can be recomputed on the
        retained images.
  \item \textbf{Which radial distortion order to adopt.} Adding higher-order
        radial coefficients enlarges the parameter space, and the
        calibration-set residual can only decrease. This decrease is an
        inherent property of the least-squares formulation and cannot
        determine whether the added coefficients improve prediction on
        retained images.
\end{itemize}

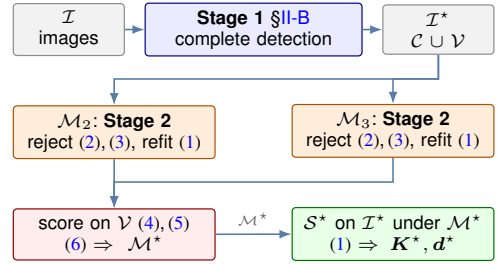
\begin{figure}[t]
\centering
\begin{tikzpicture}[node distance=2.2mm]
  \node[io, text width=13mm] (raw) {$\mathcal{I}$\\ images};
  \node[s1, right=2.6mm of raw, text width=27mm] (adm)
       {\textbf{Stage 1} \S\ref{sec:sel_a}\\ complete detection};
  \node[io, right=2.6mm of adm, text width=13mm] (pool)
       {$\mathcal{I}^\star$\\ $\mathcal{C}\cup\mathcal{V}$};
  \node[s2, text width=25mm, anchor=north west] (m2)
       at ([yshift=-6.5mm]raw.south west)
       {$\Mtwo$: \textbf{Stage 2}\\ reject \eqref{eq:tau},\,\eqref{eq:iterate},
        refit \eqref{eq:calib}};
  \node[s2, text width=25mm, anchor=north east] (m3)
       at ([yshift=-6.5mm]pool.south east)
       {$\Mthree$: \textbf{Stage 2}\\ reject \eqref{eq:tau},\,\eqref{eq:iterate},
        refit \eqref{eq:calib}};
  \node[sel, text width=25mm, anchor=north west] (val)
       at ([yshift=-6.5mm]m2.south west)
       {score on $\mathcal{V}$ \eqref{eq:valpose},\,\eqref{eq:valerr}\\
        \eqref{eq:criterion} $\Rightarrow\mathcal{M}^\star$};
  \node[res, text width=25mm, anchor=north east] (fin)
       at ([yshift=-6.5mm]m3.south east)
       {$\mathcal{S}^\star$ on $\mathcal{I}^\star$ under $\mathcal{M}^\star$\\
        \eqref{eq:calib} $\Rightarrow\bm{K}^\star,\bm{d}^\star$};
  \draw[fl] (raw) -- (adm);
  \draw[fl] (adm) -- (pool);
  \draw[fl] (pool.south) -- ++(0,-3.2mm) -| (m3.north);
  \draw[fl] (pool.south) -- ++(0,-3.2mm) -| (m2.north);
  \draw[fl] (m2.south) -- ++(0,-3.2mm) -| (val.north);
  \draw[fl] (m3.south) -- ++(0,-3.2mm) -| (val.north);
  \draw[fl] (val) -- node[lbl, above] {$\mathcal{M}^\star$} (fin);
\end{tikzpicture}
\caption{The proposed pipeline. Stage~1 accepts images once and fixes the
split. Stage~2 then runs inside each candidate order, so an order is scored
together with the estimation set its own residual scale induces. The
candidates meet only on the held-out $\mathcal{V}$, which selects
$\mathcal{M}^\star$; Stage~2 is then rerun on all of $\mathcal{I}^\star$.}
\label{fig:overview}
\vspace{-0.3cm}
\end{figure}

The two problems are coupled through the residual. Rejecting views
by their residuals requires a threshold defined relative to the 
residual scale, which is lower under a higher-order model because more
parameters fit the same corners more closely. The retained image set
therefore differs between candidate orders, and a set fixed in advance is
consistent with neither.
Existing calibration toolboxes leave both problems to the operator. The
MATLAB Camera Calibrator~\cite{matlab_calibrator} plots per-view errors, from which images are deleted
and radial coefficients enabled manually; the ROS calibrator~\cite{ros_camera_calibration} provides no
filtering stage and fixes a low-order distortion model irrespective of the
data; mrcal requires the lens model to be
specified~\cite{mrcal}. Related automation addresses other stages of the calibration problem. Guided
acquisition proposes target placements that improve
observability~\cite{richardson2013aprilcal,peng2019calibwizard, polic2020uncertainty,hagemann2022inferring,nguyen2025generalized}, which
concerns which images to capture.
Closest to this work, a patent on planar-target calibration selects
the optimal image combination under a fixed distortion
model~\cite{patent_intrinsic}, but the distortion order itself remains
predetermined.

\subsection{Contributions}
\label{sec:contrib}

The contributions of this paper are threefold:
\begin{itemize}
  \setlength{\itemsep}{1pt}\setlength{\parskip}{0pt}\setlength{\leftmargini}{1.1em}
  \item We present an automatic \textbf{image-set selection} method
        (Section~\ref{sec:selection}) for intrinsic calibration that
        iteratively rejects views whose residuals exceed a median-scaled
        threshold under each distortion order, so that the
        retained set is consistent with the residual scale.
  \item We introduce a \textbf{distortion-order selection} criterion
        (Section~\ref{sec:modelorder}) that evaluates each candidate on
        held-out images with the intrinsics and distortion fixed, avoiding
        the use of calibration-set residuals to assess model adequacy.
  \item We integrate both steps into an \textbf{interactive calibration
        tool} (Section~\ref{sec:software}) that supports full-pipeline
        data inspection and parameter estimation
        (Fig.~\ref{fig:overview}).
\end{itemize}

\section{Automatic Image Selection}
\label{sec:selection}

\subsection{Calibration Formulation}
\label{sec:notation}
Let $\mathcal{I}=\{1,\dots,N\}$ index the acquired images, with image $i$
contributing $N_i$ calibration points $\bm{u}_{ij}$ matched to target points
$\bm{P}_j$. Given intrinsics $\bm{K}$, distortion $\bm{d}$ and board pose
$\bm{T}_i\in\mathrm{SE}(3)$, the projection
$\pi(\bm{K},\bm{d},\bm{T}_i,\bm{P}_j)$ applies the Brown-Conrady model,
mapping a normalized coordinate $\bm{x}$ with $\rho=\|\bm{x}\|_2$ to
$\bm{x}(1 + k_1 \rho^2 + k_2 \rho^4 + k_3 \rho^6) +
\bm{t}(\bm{x};p_1,p_2)$. Calibration over
$\mathcal{S}\subseteq\mathcal{I}$ minimizes
\begin{equation}
\label{eq:calib}
\min_{\bm{K},\bm{d},\{\bm{T}_i\}}
\sum_{i\in\mathcal{S}}\sum_{j=1}^{N_i}
\bigl\|\bm{u}_{ij}-\pi(\bm{K},\bm{d},\bm{T}_i,\bm{P}_j)\bigr\|_2^2 ,
\end{equation}
with per-view mean residual $\meanerr_i = N_i^{-1}\sum_j\|\bm{u}_{ij}-
\pi(\bm{K},\bm{d},\bm{T}_i,\bm{P}_j)\|_2$. The calibration-set error is
evaluated on the images to estimate $\bm{K},\bm{d}$. The held-out error
is evaluated on excluded images with $\bm{K},\bm{d}$ fixed and only
$\bm{T}_i$ re-estimated.

\subsection{Stage 1: Detection-Based Acceptance}
\label{sec:sel_a}
Sub-pixel corner extraction has been improved by dedicated
detectors~\cite{placht2014rochade,duda2018accurate} and by target designs
that raise the precision of each
junction~\cite{ha2017deltille}. In this work, the output of such a detector
serves as the acceptance criterion of Stage~1: an image is admitted to the
estimation only if the extractor~\cite{duda2018accurate} recovers the
complete inner corner grid, which also excludes partial detections that
associate different sub-grids across images. This criterion is operational:
the extractor fails predominantly under motion blur and at extreme
incidence, precisely the conditions under which corner extraction bias
propagates into the shared intrinsics. The
acceptance criterion applies only when at least
$\max(N_{\mathrm{stop}},\, N/2)$ images are retained; fewer indicates a
problem with the target or the acquisition rather than with individual
images.

\subsection{Stage 2: Iterative Robust Rejection}
\label{sec:sel_b}
While Stage~1 filters images based on detection completeness, it does not
identify images whose corners are extracted successfully but are
geometrically inconsistent with the rest of the set. Such inconsistencies
become visible only after an initial calibration, through the per-view
residuals $\meanerr_i$, and removing them is the task of Stage~2. Following
prior work on planar-target calibration~\cite{patent_intrinsic}, we
iteratively reject images whose residual exceeds a threshold proportional
to the median. Specifically, let $\mathcal{I}^{(t)}$ denote the retained set
at iteration $t$ and
$\medres^{(t)} = \mathrm{med}\{\meanerr_i\}_{i\in\mathcal{I}^{(t)}}$.
Images are rejected by
\begin{align}
\label{eq:tau}
\tau^{(t)} &= \kappa\,\medres^{(t)}, \qquad \kappa = 2, \\
\label{eq:iterate}
\mathcal{I}^{(t+1)} &= \bigl\{\, i\in\mathcal{I}^{(t)}
   \;\bigm|\; \meanerr_i^{(t)} \le \tau^{(t)} \,\bigr\} ,
\end{align}
after which \eqref{eq:calib} is re-solved on $\mathcal{I}^{(t+1)}$.
Because $\medres^{(t)}$ is recomputed at each iteration, the threshold
tightens as outlier views are removed, and also adapts to the candidate
distortion order, since a higher-order model yields a lower residual scale.
The robust scale estimate
$\sighat = 1.4826 \cdot \mathrm{med}_{i}\bigl(|\meanerr_i -
\medres|\bigr)$~\cite{rousseeuw1993mad} is evaluated as an alternative in
Section~\ref{sec:exp_sel}. The iteration terminates when no image exceeds
$\tau^{(t)}$, after at most $T_{\max}=3$ rounds, or when fewer than
$N_{\mathrm{stop}}=8$ images remain. Crucially, this procedure runs
independently under each candidate order, producing a model-conditioned
estimation set $\mathcal{S}_\mathcal{M}$.

\section{Distortion-Model Selection}
\label{sec:modelorder}

\subsection{Candidate Models and Subsets}
\label{sec:split}
The comparison is restricted to the radial order:
$\Mtwo=\{k_1,k_2,p_1,p_2\}$ versus $\Mthree=\{k_1,k_2,k_3,p_1,p_2\}$,
the default choice in mainstream toolboxes. Rational and fisheye
families~\cite{kannala2006generic}, as well as formulations that unify
several models~\cite{lochman2021babelcalib}, are beyond the scope of this
work but compatible with the selection criterion below.

The set $\mathcal{I}^\star$ accepted by Stage~1 is split into an estimation
subset $\mathcal{C}$ and a validation subset $\mathcal{V}$ by assigning
every fifth image in capture order to $\mathcal{V}$, distributing the
validation views over the acquisition sequence and making the split
deterministic. This split is applied only when
$|\mathcal{I}^\star|\ge N_{\min}=15$; below this threshold $\Mtwo$ is
retained. Importantly, $\mathcal{V}$ must be drawn from
$\mathcal{I}^\star$ rather than from the images rejected at Stage~1: a
validation subset with higher corner noise than the estimation subset would
reflect its own extraction error rather than the distortion model, biasing
the comparison toward the lower order (Section~\ref{sec:exp_order}).

\subsection{Pose-Only Validation and Criterion}
For each candidate $\mathcal{M}$, \eqref{eq:calib} is solved over
$\mathcal{C}$ with the rejection of~\eqref{eq:iterate} applied internally,
so that Stage~2 operates under the candidate model itself, yielding
$(\bm{K}_\mathcal{M},\bm{d}_\mathcal{M})$. The estimated intrinsics and
distortion are then fixed, and only the board pose is re-estimated on each
validation image by solving
\begin{equation}
\label{eq:valpose}
\hat{\bm{T}}_i^\mathcal{M} = \arg\min_{\bm{T}\in\mathrm{SE}(3)}
\sum_{j=1}^{N_i}
\bigl\|\bm{u}_{ij}-\pi(\bm{K}_\mathcal{M},\bm{d}_\mathcal{M},\bm{T},
\bm{P}_j)\bigr\|_2^2 ,
\end{equation}
for each $i\in\mathcal{V}$, which is a perspective-$n$-point problem with
six degrees of freedom and $2N_i$ measurements. Substituting
$\hat{\bm{T}}_i^\mathcal{M}$ into $\meanerr_i$ yields per-view validation
errors, summarized as
\begin{equation}
\label{eq:valerr}
\eval(\mathcal{M})
 = \frac{1}{|\mathcal{V}|}\sum_{i\in\mathcal{V}}\meanerr_i^\mathcal{M},
\quad
\emaxv(\mathcal{M}) = \max_{i\in\mathcal{V}}
 \meanerr_i^\mathcal{M}.
\end{equation}
By refitting only the six pose parameters with the intrinsics and
distortion fixed, each candidate is evaluated on views that did not
contribute to its estimation. Each order is compared together with the
estimation set that its own residual scale induces, and the common
validation subset $\mathcal{V}$ ensures that the comparison is controlled.
Formally, $\Mthree$ is adopted if and only if
\begin{equation}
\label{eq:criterion}
\eval(\Mthree) \le \alpha\,\eval(\Mtwo)
\;\;\text{and}\;\;
\emaxv(\Mthree) \le \beta\, \emaxv(\Mtwo),
\end{equation}
with $\alpha=1$ and $\beta=1$, both evaluated on views excluded from the
intrinsic estimation. That is, the higher order is adopted only when it
degrades neither the mean nor the worst-case validation error, requiring
that the added coefficients be supported by independent observations.
Parity rather than a strict margin is used to avoid penalizing differences
within the noise level of $\eval$; the sensitivity to $\alpha$ is examined
in Section~\ref{sec:exp_order}.
Once $\mathcal{M}^\star$ is determined, the rejection of
Section~\ref{sec:sel_b} is rerun on the full accepted pool
$\mathcal{I}^\star$ under $\mathcal{M}^\star$, producing the final
estimation set $\mathcal{S}^\star\subseteq\mathcal{I}^\star$ from which
$(\bm{K}^\star,\bm{d}^\star)$ are obtained. The validation views enter the
final estimate at this stage, and the views rejected in this pass are
recorded alongside the rest.

\begin{figure}[t]
\centering
\includegraphics[width=0.92\linewidth]{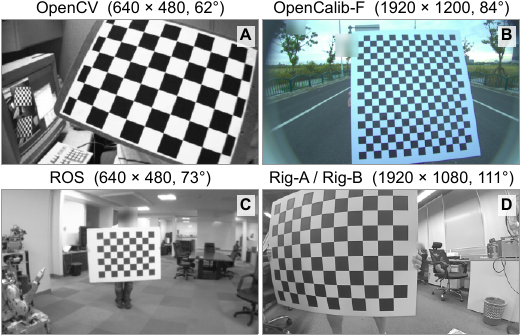}
\caption{The four camera models evaluated, $62^\circ$ to $111^\circ$ diagonal
field of view: (A)~OpenCV, $9\times6$ board; (B)~OpenCalib-F, $17\times15$,
the only outdoor scene; (C)~ROS, $8\times6$; (D)~Rig-A/Rig-B, $11\times8$,
captured on our own rig.}
\label{fig:datasets}
\vspace{-0.3cm}
\end{figure}


\section{Interactive Calibration Tool}
\label{sec:software}

The image filtering and distortion-order selection of
Sections~\ref{sec:selection} and~\ref{sec:modelorder} introduce
data-dependent decisions into the calibration procedure. To make these
decisions transparent and reproducible, we implement the pipeline as an
interactive offline tool whose workflow consists of three steps: importing
an image folder, specifying the target geometry, and running the
calibration. The tool displays the filtering status of every image and the
distribution of calibration points, so that both decisions can be
inspected within the interface (Supplement~\ref{app:arch},
Fig.~\ref{fig:gui}). The algorithmic core is independent of the interface
and can be invoked from scripts; all results in Section~\ref{sec:exp} are
produced in this way. In addition to the estimated parameters, the tool
records the calibration provenance: the images retained and rejected at
each stage with their residuals, the subsets $\mathcal{C}$ and
$\mathcal{V}$, the four quantities of~\eqref{eq:criterion} with the
selected order, and the number of observations in the four $20\%$ corner
zones of the image.

\section{Experiments}
\label{sec:exp}

\subsection{Experimental Setup}
\label{sec:exp_setup}
\textbf{Data.} We evaluate on own cameras images and five public
datasets, covering four camera models that span $62^\circ$ to $111^\circ$
diagonal field of view and $0.3$ to $2.3$\,Mpx
(Fig.~\ref{fig:datasets}). \textbf{Rig-A}, the primary dataset, contains
39 images captured hand-held on a mobile mapping rig.
\textbf{Rig-B} is a second camera of the same model, captured in two
independent sessions. The public datasets are OpenCalib's front
camera~\cite{yan2022opencalib} and the stereo pairs distributed with ROS
and OpenCV. Chessboard specifications (inner corners, square size): Rig
$11\times8$, 45\,mm; OpenCalib-F $17\times15$, 50\,mm; ROS $8\times6$,
108\,mm; OpenCV $9\times6$, 30\,mm.

\textbf{Baselines.} We compare against two toolboxes run with their
default settings. The first is the \texttt{MonoCalibrator} of the ROS
\texttt{camera\_calibration} package~\cite{ros_camera_calibration} (ROS-calib), which
uses its own detector, sub-pixel refinement and frame acceptance at
\texttt{-k\,2} ($\Mtwo$). The second is
mrcal~\cite{mrcal}, which includes outlier rejection and a board-flex
model; we apply it to the corners extracted in
Section~\ref{sec:sel_a} at its documented 8-coefficient rational model,
with the 4- and 5-coefficient variants included for the order comparison.
Additionally, the proposed pipeline with both selections disabled, no
image filtering and the order fixed at $\Mtwo$, serves as an ablation
baseline. For the public datasets, we adopt a rotating four-fold
holdout protocol: both selection stages, the order criterion and the final
fit are computed within the calibration fold, ensuring that no held-out image
informs the selection.

\textbf{Metrics.} In the absence of ground-truth intrinsics, reprojection
error is the standard image-space
metric~\cite{schops2020generic,nguyen2025generalized}: the calibration-set
residual serves as a fit diagnostic, and the held-out error as the
evaluation criterion. We report the mean corner error
$\meanerr=\operatorname{mean}_{ij}\|\bm{u}_{ij}-\hat{\bm{u}}_{ij}\|_2$,
following the MATLAB convention. For Rig-A,
$\mathcal{V}$ consists of the same six images throughout all experiments.
With both selection stages disabled, the estimator reproduces the ROS
calibrator to within $2.1$\,px on Rig-B, confirming that the improvements
reported below originate from the selection method rather than from
differences in the solver.

\begin{figure}[t]
\centering
\includegraphics[width=\linewidth]{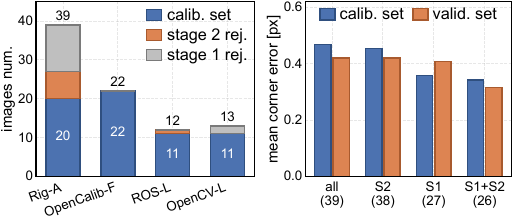}
\caption{Image selection. Left: images kept and rejected per stage; right
stereo cameras omitted (identical to left). Right: calibration and validation
mean corner error [px] of four Rig-A sets, $\Mtwo$.}
\label{fig:selection}
\vspace{-0.3cm}
\end{figure}

\begin{figure}[!t]
\centering
\includegraphics[width=\linewidth]{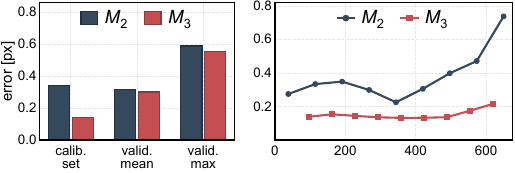}
\caption{Distortion-order selection on Rig-A. Left: mean corner error [px] of
both candidates on the calibration set and the validation views. Right:
per-view residual against the radial distance $r$ from the principal point,
over the observed range.}
\label{fig:order}
\vspace{-0.3cm}
\end{figure}

\subsection{Image Selection Analysis}
\label{sec:exp_sel}
This experiment evaluates the image filtering of
Section~\ref{sec:selection}.

Fig.~\ref{fig:selection} (left) summarizes the filtering outcome across
all datasets. On Rig-A, Stage~1 rejects 12 of 39 images and Stage~2
rejects a further 7 under the final selected order; on OpenCalib-F, all 22 images pass both stages; on
the remaining datasets, one to two images are rejected. The median
per-view error of rejected images on Rig-A is $4.7$ times that of the
accepted ones, confirming that the filtering targets views most affected
by systematic corner extraction error.
Fig.~\ref{fig:selection} (right) compares four configurations on Rig-A
under $\Mtwo$, evaluated on the same six held-out images. Applying both
stages reduces the held-out error from 0.419 to 0.316\,px ($25\%$),
substantially more than the calibration-set reduction from 0.467 to
0.342\,px. This asymmetry is expected: removing systematically biased views
benefits the shared intrinsics more than it benefits the calibration-set
residual, which is partially absorbed by the per-view pose parameters.
The two stages serve complementary roles: without Stage~1, Stage~2
rejects only one image and the held-out error remains at 0.419\,px,
because high-residual views inflate the median on which the threshold
depends. The median-scaled threshold of~\eqref{eq:tau} is preferable to
$\sighat$, which either retains all images or over-rejects
(Supplement~\ref{app:perview}).

\subsection{Distortion-Model Selection Analysis}
\label{sec:exp_order}
This experiment evaluates the order selection criterion of
Section~\ref{sec:modelorder}.
Fig.~\ref{fig:order} compares $\Mtwo$ and $\Mthree$ on Rig-A. The
calibration-set error drops from 0.342\,px under $\Mtwo$ to 0.143\,px
under $\Mthree$, but this reduction conflates two effects: on the 26
images retained under $\Mtwo$, the additional coefficient alone reduces
the error from 0.342 to 0.252\,px; the
remainder is attributable to the six further images rejected as the
residual scale tightens under $\Mthree$. Neither quantity can determine
the order, as both are derived from the data that produced the fit. On
the held-out views, the difference is 0.316\,px versus 0.299\,px,
satisfying both conditions of~\eqref{eq:criterion} ($\emaxv$:
$0.554 \le 0.590$). The order selection also shifts $f_x$ by $11$\,px to
$661.73$\,px. Repeating the split at all five offsets yields a
consistently lower $\eval$ under $\Mthree$ (mean difference
$-0.073\pm0.034$\,px); the criterion adopts $\Mthree$ at four of five
offsets. Setting $\alpha=0.98$ instead of
parity would reject $\Mthree$ on two OpenCalib-F folds whose improvement
is only $1.2$ and $1.3\%$, confirming that parity is appropriate at this
noise level.

The order criterion requires at least
$N_{\min}$ images per estimation fold, satisfied only by the two
wider-field datasets. On OpenCalib-F, neither stage rejects any image, so
the improvement is due to order selection alone: $\Mthree$ is selected on
three of four folds, reducing the held-out error from 0.230 to
0.226\,px; on the fourth fold the worst validation view degrades and
$\Mtwo$ is retained. The four narrower-field datasets fall below
$N_{\min}$; only image filtering contributes, reducing the held-out mean
by 0.6 to 2.6\%.
In Fig.~\ref{fig:order} (right), the per-view
residual under $\Mtwo$ increases from 0.23 to 0.74\,px toward the image
border while under $\Mthree$ it remains near 0.15\,px, indicating
under-modelling by $\Mtwo$. This holds only over the observed range:
the calibration points extend to
$r=841$\,px versus $r=1126$\,px at the corner, where the two radial
scale factors diverge by $21.2\%$.

\begin{table}[t]
\centering
\caption{Held-out mean corner error [px]~$\downarrow$ of each method on the
five public datasets.}
\label{tab:cross}
\footnotesize
\renewcommand{\arraystretch}{1.3}
\begin{threeparttable}
\setlength{\tabcolsep}{1.6pt}
\begin{tabular}{|l||c||c|c||c|c|}
\hline
\multirow{2}{*}{\textbf{Method}} & \multirow{2}{*}{\textbf{OpenCalib-F}}
 & \multicolumn{2}{c||}{\textbf{ROS}} & \multicolumn{2}{c|}{\textbf{OpenCV}} \\
\cline{3-6}
 & & L & R & L & R \\
\hline
ROS-calib
 & \cellcolor{tabbest!30}0.232(18)  & \cellcolor{tabbest!30}0.168(41)
 & \cellcolor{tabbest!30}0.157(34)  & \cellcolor{tabbest!10}0.601(84)
 & \cellcolor{tabbest!10}0.665(101) \\
mrcal
 & \cellcolor{tabbest!10}0.387(104) & \cellcolor{tabbest!10}0.231(77)
 & \cellcolor{tabbest!10}0.165(35)  & \cellcolor{tabbest!30}0.243(50)
 & \cellcolor{tabbest!30}0.271(50) \\
\hline
ours w/o sel.\
 & \cellcolor{tabbest!60}0.230(18)  & \cellcolor{tabbest!60}0.167(40)
 & \cellcolor{tabbest!60}0.151(33)  & \cellcolor{tabbest!60}0.234(51)
 & \cellcolor{tabbest!60}0.246(55) \\
ours
 & \cellcolor{tabbest}\textbf{0.226}(21) & \cellcolor{tabbest}\textbf{0.165}(40)
 & \cellcolor{tabbest}\textbf{0.148}(34) & \cellcolor{tabbest}\textbf{0.233}(49)
 & \cellcolor{tabbest}\textbf{0.240}(58) \\
\hline
\end{tabular}
\begin{tablenotes}[flushleft]
\item \scriptsize\textbf{Note:} parentheses give the fold-to-fold standard
deviation in the last digit; lower is better ($\downarrow$): within each
column, a darker cell marks a lower error, \textbf{bold} the best;
``w/o sel.'' disables selection.
\end{tablenotes}
\end{threeparttable}
\end{table}

\subsection{Comparison with Existing Methods}
\label{sec:exp_cross}
Table~\ref{tab:cross} reports the held-out error of each method on the
five public datasets. The proposed pipeline achieves the lowest held-out
mean on all five datasets. Since the fold-to-fold standard deviation
exceeds the difference between the full pipeline and the ablation,
the paired per-fold difference is more discriminative: on
OpenCalib-F it ranges from $-6.3\%$ to $0\%$ across the four folds, with
the three folds that adopt $\Mthree$ showing improvement and the
remaining fold unchanged.
The mrcal results corroborate the need for order selection. At 8
coefficients, mrcal achieves the lowest calibration-set residual of any
configuration, yet its held-out error exceeds that at 4 coefficients by
7.6 to 63.3\% across all five datasets, a direct instance of the
overfitting described in Section~\ref{sec:challenges}. On OpenCalib-F,
the 5-coefficient model outperforms the 4-coefficient one, consistent
with the criterion selecting $\Mthree$. On Rig-A, the proposed pipeline
yields 0.143\,px on the retained images, compared with 0.151\,px for
mrcal and 0.299\,px for ROS-calib.


\section{Conclusion}
\label{sec:conclusion}

This paper presented a fully automatic camera intrinsic calibration
pipeline that determines both the image set and the radial distortion
order from the collected data, without relying on the calibration-set
residual for either decision. Experiments on seven datasets across four
camera models show that image filtering reduces the held-out error by up
to 25\%, with order selection providing further improvement. Future work
includes extending the candidate set to rational and fisheye distortion
families and incorporating geometric coverage constraints into the
filtering criterion.

\section*{Acknowledgment}
Generative AI tools were used to polish the wording of this manuscript and
to assist the development of the graphical user interface of the
calibration tool.

\clearpage
\bibliographystyle{IEEEtran}
\bibliography{refs}

\begin{thebibliography}{10}
\providecommand{\url}[1]{#1}
\csname url@rmstyle\endcsname
\providecommand{\newblock}{\relax}
\providecommand{\bibinfo}[2]{#2}
\providecommand\BIBentrySTDinterwordspacing{\spaceskip=0pt\relax}
\providecommand\BIBentryALTinterwordstretchfactor{4}
\providecommand\BIBentryALTinterwordspacing{\spaceskip=\fontdimen2\font plus
\BIBentryALTinterwordstretchfactor\fontdimen3\font minus
  \fontdimen4\font\relax}
\providecommand\BIBforeignlanguage[2]{{%
\expandafter\ifx\csname l@#1\endcsname\relax
\typeout{** WARNING: IEEEtran.bst: No hyphenation pattern has been}%
\typeout{** loaded for the language `#1'. Using the pattern for}%
\typeout{** the default language instead.}%
\else
\language=\csname l@#1\endcsname
\fi
#2}}

\bibitem{hu2024paloc}
X.~Hu \emph{et~al.}, ``{PALoc}: Advancing {SLAM} benchmarking with
  prior-assisted 6-{DoF} trajectory generation and uncertainty estimation,''
  \emph{IEEE/ASME Transactions on Mechatronics}, vol.~29, no.~6, pp.
  4297--4308, 2024.

\bibitem{schops2020generic}
T.~Sch\"{o}ps \emph{et~al.}, ``Why having 10,000 parameters in your camera
  model is better than twelve,'' in \emph{IEEE/CVF Conf. Comput. Vis. Pattern
  Recognit. (CVPR)}, 2020, pp. 2535--2544.

\bibitem{hu2025mapeval}
X.~Hu \emph{et~al.}, ``{MapEval}: Towards unified, robust and efficient {SLAM}
  map evaluation framework,'' \emph{IEEE Robot. Autom. Lett.}, vol.~10, no.~5,
  pp. 4228--4235, 2025.

\bibitem{zhang2000flexible}
Z.~Zhang, ``A flexible new technique for camera calibration,'' \emph{IEEE
  Trans. Pattern Anal. Mach. Intell.}, vol.~22, no.~11, pp. 1330--1334, 2000.

\bibitem{brown1971close}
D.~C. Brown, ``Close-range camera calibration,'' \emph{Photogramm. Eng.},
  vol.~37, no.~8, pp. 855--866, 1971.

\bibitem{matlab_calibrator}
{The MathWorks, Inc.}, ``Using the single camera calibrator app,'' Comput. Vis.
  Toolbox Doc.,
  \url{https://www.mathworks.com/help/vision/ug/using-the-single-camera-calibrator-app.html},
  2026, accessed: 2026-08-16.

\bibitem{ros_camera_calibration}
{ROS Perception}, ``{camera\_calibration},'' ROS Wiki,
  \url{http://wiki.ros.org/camera_calibration}, accessed: 2026-08-16.

\bibitem{mrcal}
D.~Kogan, ``{mrcal}: Camera-modeling toolkit,''
  \url{https://mrcal.secretsauce.net}, 2026, version 2.5.2, accessed:
  2026-08-16.

\bibitem{tang2017precision}
Z.~Tang \emph{et~al.}, ``A precision analysis of camera distortion models,''
  \emph{IEEE Trans. Image Process.}, vol.~26, no.~6, pp. 2694--2704, 2017.

\bibitem{richardson2013aprilcal}
A.~Richardson \emph{et~al.}, ``{AprilCal}: Assisted and repeatable camera
  calibration,'' in \emph{IEEE/RSJ Int. Conf. Intell. Robots Syst. (IROS)},
  2013, pp. 1814--1821.

\bibitem{peng2019calibwizard}
S.~Peng and P.~Sturm, ``Calibration wizard: A guidance system for camera
  calibration based on modelling geometric and corner uncertainty,'' in
  \emph{IEEE/CVF Int. Conf. Comput. Vis. (ICCV)}, 2019, pp. 1497--1505.

\bibitem{polic2020uncertainty}
M.~Polic \emph{et~al.}, ``Uncertainty based camera model selection,'' in
  \emph{IEEE/CVF Conf. Comput. Vis. Pattern Recognit. (CVPR)}, 2020, pp.
  5991--6000.

\bibitem{hagemann2022inferring}
A.~Hagemann \emph{et~al.}, ``Inferring bias and uncertainty in camera
  calibration,'' \emph{Int. J. Comput. Vis.}, vol. 130, no.~1, pp. 17--32,
  2022.

\bibitem{nguyen2025generalized}
C.~Q. Nguyen and S.~Choi, ``Generalized camera calibration: Camera model
  selection and calibration with effective image sampling,'' \emph{IEEE Sensors
  J.}, vol.~25, no.~15, pp. 29\,124--29\,140, 2025.

\bibitem{patent_intrinsic}
{U.S. Patent}, ``Method and apparatus for automatic intrinsic camera
  calibration using images of a planar calibration pattern,'' U.S. Patents
  10{,}269{,}140; 10{,}380{,}766; 10{,}706{,}588; 12{,}243{,}269, 2019--2025.

\bibitem{placht2014rochade}
S.~Placht \emph{et~al.}, ``{ROCHADE}: Robust checkerboard advanced detection
  for camera calibration,'' in \emph{Eur. Conf. Comput. Vis. (ECCV)}, 2014, pp.
  766--779.

\bibitem{duda2018accurate}
A.~Duda and U.~Frese, ``Accurate detection and localization of checkerboard
  corners for calibration,'' in \emph{Brit. Mach. Vis. Conf. (BMVC)}, 2018.

\bibitem{ha2017deltille}
H.~Ha \emph{et~al.}, ``Deltille grids for geometric camera calibration,'' in
  \emph{IEEE/CVF Int. Conf. Comput. Vis. (ICCV)}, 2017, pp. 5354--5362.

\bibitem{rousseeuw1993mad}
P.~J. Rousseeuw and C.~Croux, ``Alternatives to the median absolute
  deviation,'' \emph{J. Amer. Statist. Assoc.}, vol.~88, no. 424, pp.
  1273--1283, 1993.

\bibitem{kannala2006generic}
J.~Kannala and S.~S. Brandt, ``A generic camera model and calibration method
  for conventional, wide-angle, and fish-eye lenses,'' \emph{IEEE Trans.
  Pattern Anal. Mach. Intell.}, vol.~28, no.~8, pp. 1335--1340, 2006.

\bibitem{lochman2021babelcalib}
Y.~Lochman \emph{et~al.}, ``{BabelCalib}: A universal approach to calibrating
  central cameras,'' in \emph{IEEE/CVF Int. Conf. Comput. Vis. (ICCV)}, 2021,
  pp. 15\,233--15\,242.

\bibitem{yan2022opencalib}
G.~Yan \emph{et~al.}, ``{OpenCalib}: A multi-sensor calibration toolbox for
  autonomous driving,'' \emph{Softw. Impacts}, vol.~14, p. 100393, 2022.

\end{thebibliography}

\twocolumn[{%
\begin{center}
  \vspace{4mm}
  {\LARGE\bf Supplementary Material}\\[2.5mm]
\end{center}
\vspace{5mm}
}]
\setcounter{section}{0}
\setcounter{subsection}{0}
\renewcommand{\thesubsection}{\Alph{subsection}}


This supplementary material presents the complete pipeline and the
per-view filtering statistics~(\ref{app:alg}), and describes the software
implementation~(\ref{app:arch}). The tool, the data and the scripts that
reproduce every number in the paper will be released; the repository is
withheld during review.

\begin{algorithm}[t]
\small
\caption{Image and distortion-order selection}
\label{alg:pipeline}
\begin{algorithmic}[1]
\REQUIRE $\mathcal{I}$, target geometry,
         $\kappa,\alpha,\beta,N_{\min},N_{\mathrm{stop}},T_{\max}$
\STATE $\mathcal{A}\leftarrow\{\,i\in\mathcal{I}\mid$ complete
       detection by~\cite{duda2018accurate} $\}$
       \COMMENT{\S\ref{sec:sel_a}}
\STATE $\mathcal{I}^\star\leftarrow\mathcal{A}$ if
       $|\mathcal{A}|\ge\max(N_{\mathrm{stop}},\,|\mathcal{I}|/2)$,
       else all detected images
\IF{$|\mathcal{I}^\star| \ge N_{\min}$}
  \STATE $\mathcal{V}\leftarrow\mathcal{I}^\star_{0::5}$,\;
         $\mathcal{C}\leftarrow\mathcal{I}^\star\setminus\mathcal{V}$
         \COMMENT{\S\ref{sec:split}}
  \FORALL{$\mathcal{M}\in\{\Mtwo,\Mthree\}$}
    \STATE $(\bm{K}_\mathcal{M},\bm{d}_\mathcal{M})\leftarrow
           \textsc{Select}(\mathcal{C},\mathcal{M})$
    \STATE $(\eval,\emaxv)(\mathcal{M})\leftarrow$
           \eqref{eq:valpose},\,\eqref{eq:valerr} on $\mathcal{V}$
  \ENDFOR
  \STATE $\mathcal{M}^\star\leftarrow\Mthree$ if \eqref{eq:criterion},
         else $\Mtwo$
\ELSE
  \STATE $\mathcal{M}^\star\leftarrow\Mtwo$
\ENDIF
\RETURN $\textsc{Select}(\mathcal{I}^\star,\mathcal{M}^\star)$ and the
        calibration record
\STATE
\STATE \textbf{function} $\textsc{Select}(\mathcal{S},\mathcal{M})$
\STATE \quad \textbf{repeat} solve \eqref{eq:calib} on $\mathcal{S}$
       under $\mathcal{M}$; $\mathcal{S}'\leftarrow$ \eqref{eq:tau},
       \eqref{eq:iterate}; \textbf{if} $\mathcal{S}'=\mathcal{S}$ or
       $|\mathcal{S}'|<N_{\mathrm{stop}}$ \textbf{break else}
       $\mathcal{S}\leftarrow\mathcal{S}'$
\STATE \quad \textbf{until} $T_{\max}$ rounds; \textbf{return}
       $(\bm{K},\bm{d})$ on $\mathcal{S}$
\end{algorithmic}
\end{algorithm}

\begin{figure}[!t]
\centering
\includegraphics[width=\linewidth]{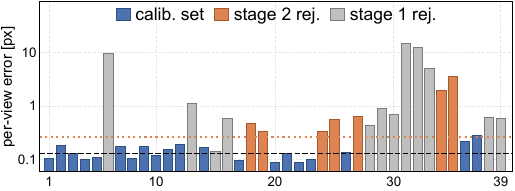}
\caption{Per-view residual of all 39 Rig-A images, under the final model and colored by selection result. Dashed: median
of the calibration set; dotted: rejection threshold $\tau=2\medres$.}
\label{fig:perview}
\vspace{-0.3cm}
\end{figure}

\subsection{Complete Pipeline}
\label{app:alg}
Algorithm~\ref{alg:pipeline} presents the complete pipeline, including
the fallback criterion that retains $\Mtwo$ when the accepted set is too
small to split.
\label{app:perview}
Fig.~\ref{fig:perview} shows the per-view residuals of all 39 Rig-A
images under the final model, alongside the filtering outcome.
Replacing the median-scaled threshold of~\eqref{eq:tau} by
$\medres+\lambda\sighat$ leaves the retained set unchanged for
$\lambda\ge3$; for $\lambda\le2$, only 9 to 13 of the 21 estimation
images are retained and the held-out error degrades to 0.409\,px. The
twelve images rejected by Stage~1 have a median residual $4.7$ times
that of the accepted ones, and the seven rejected by Stage~2 are those
exceeding $\tau=\kappa\medres$ after the residual scale has been
recomputed on the set that Stage~1 retained.




\subsection{Software Implementation}
\label{app:arch}
\begin{figure}[!t]
  \centering
  \includegraphics[width=\linewidth]{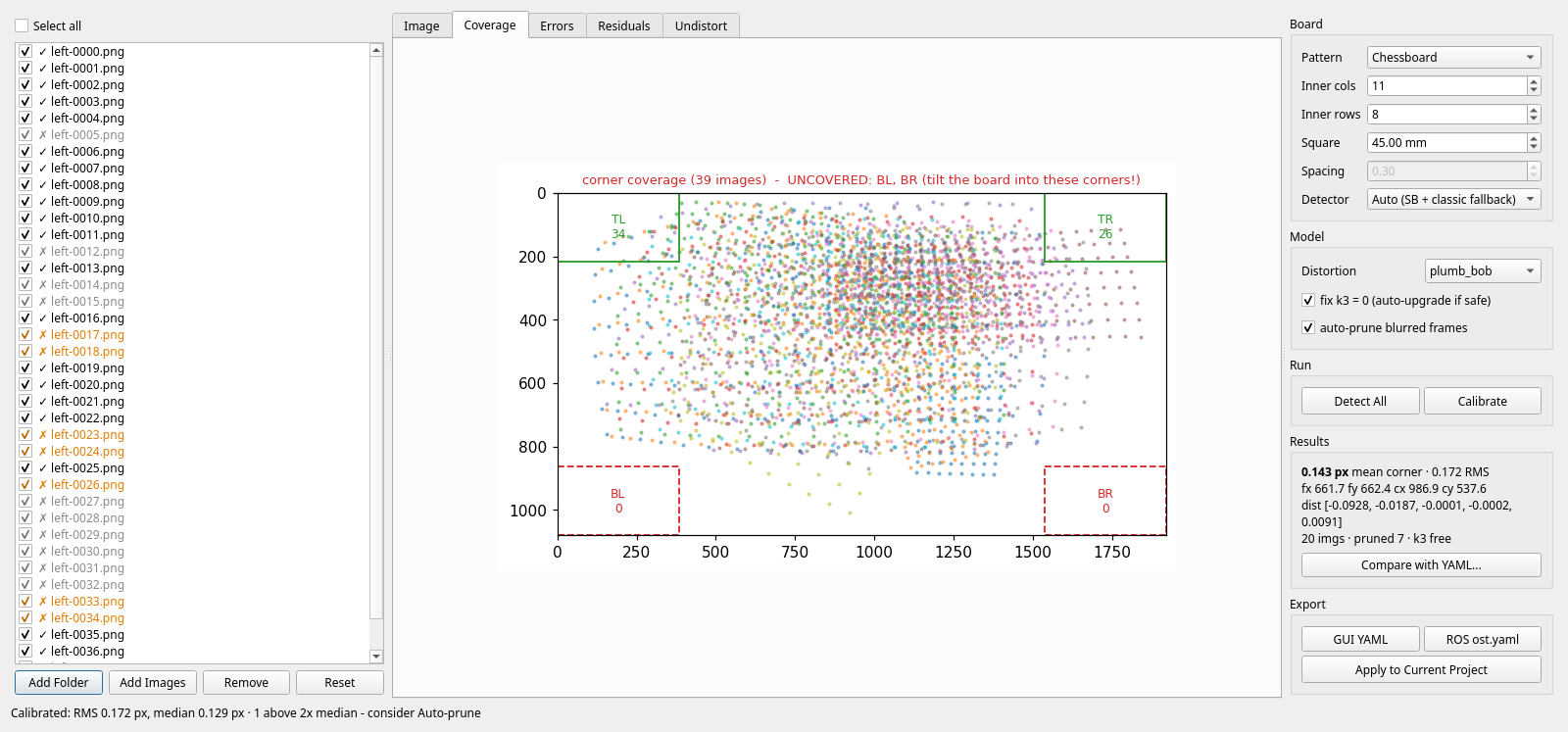}
  \caption{Interface on Rig-A after calibration. Left: image list, with
  images rejected at Stage~1 greyed out and those rejected at Stage~2
  marked in orange. Center: calibration points and corner-zone counts.
  Right: board and model settings with the estimated parameters. The
  status bar reports the residual and the number of views exceeding
  $\kappa\medres$.}
  \label{fig:gui}
\end{figure}

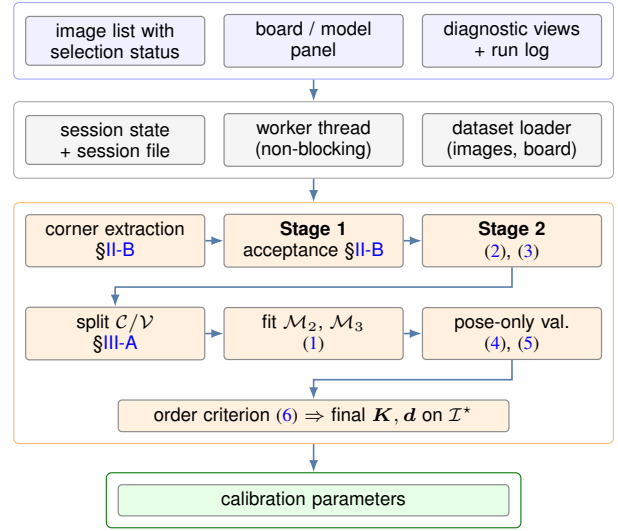
\begin{figure}[!t]
\centering
\begin{tikzpicture}[node distance=2mm and 2.4mm]
  \node[sw, fill=blue!6] (ui1) {image list with\\selection status};
  \node[sw, fill=blue!6, right=of ui1] (ui2) {board / model\\panel};
  \node[sw, fill=blue!6, right=of ui2] (ui3) {diagnostic views\\+ run log};
  \node[sw, fill=gray!8, below=6.5mm of ui1] (se1) {session state\\+ session file};
  \node[sw, fill=gray!8, right=of se1] (se2) {worker thread\\(non-blocking)};
  \node[sw, fill=gray!8, right=of se2] (se3) {dataset loader\\(images, board)};
  \node[sw, fill=orange!12, below=6.5mm of se1] (c1)
       {corner extraction\\\S\ref{sec:sel_a}};
  \node[sw, fill=orange!12, right=of c1] (c2)
       {\textbf{Stage 1}\\acceptance \S\ref{sec:sel_a}};
  \node[sw, fill=orange!12, right=of c2] (c3)
       {\textbf{Stage 2}\\\eqref{eq:tau}, \eqref{eq:iterate}};
  \node[sw, fill=orange!12, below=5.2mm of c1] (c5)
       {split $\mathcal{C}/\mathcal{V}$\\\S\ref{sec:split}};
  \node[sw, fill=orange!12, right=of c5] (c6)
       {fit $\Mtwo$, $\Mthree$\\\eqref{eq:calib}};
  \node[sw, fill=orange!12, right=of c6] (c7)
       {pose-only val.\\\eqref{eq:valpose}, \eqref{eq:valerr}};
  \node[sw, fill=orange!12, below=5.2mm of c6, text width=50mm] (c4)
       {order criterion \eqref{eq:criterion} $\Rightarrow$ final
        $\bm{K},\bm{d}$ on $\mathcal{I}^\star$};
  \draw[fl] (c1) -- (c2);  \draw[fl] (c2) -- (c3);
  \draw[fl] (c3.south) -- ++(0,-2.6mm) -| (c5.north);
  \draw[fl] (c5) -- (c6);  \draw[fl] (c6) -- (c7);
  \draw[fl] (c7.south) -- ++(0,-2.6mm) -| (c4.north);
  \node[sw, fill=green!10, text width=50mm] (e1)
       at ($(c4.south)+(0,-9mm)$) {calibration parameters};
  \begin{scope}[on background layer]
    \node[draw=blue!35, rounded corners=2pt, inner sep=1.5mm,
          fit=(ui1)(ui3)] (L1) {};
    \node[draw=black!35, rounded corners=2pt, inner sep=1.5mm,
          fit=(se1)(se3)] (L2) {};
    \node[draw=orange!55, rounded corners=2pt, inner sep=1.5mm,
          fit=(c1)(c3)(c5)(c7)(c4)] (L3) {};
    \node[draw=green!45!black, rounded corners=2pt, inner sep=1.5mm,
          fit=(e1)] (L4) {};
  \end{scope}
  \draw[fl] (L1) -- (L2);
  \draw[fl] (L2) -- (L3);
  \draw[fl] (L3) -- (L4);
\end{tikzpicture}
\caption{Software layers. The core reproduces a calibration; the session
and interface layers add the diagnostics and the results.}
\label{fig:arch}
\vspace{-0.3cm}
\end{figure}

Fig.~\ref{fig:arch} illustrates the four layers of the tool.
The {core} layer implements corner extraction
(Section~\ref{sec:sel_a}), both selection stages
(Sections~\ref{sec:sel_a}--\ref{sec:sel_b}), the
estimation~\eqref{eq:calib} and the order
criterion~\eqref{eq:criterion}. The {session} layer manages the
dataset and delegates computation to a worker thread. The
{interface} layer (Fig.~\ref{fig:gui}) exposes three panels: an
image list showing the selection status of every frame, a calibration-point
view with the per-zone corner counts, and a
board and model panel with the estimated parameters. Images rejected at
Stage~1 are re-detected with the OpenCV corner extractor, using a
sub-pixel refinement window of $0.35$ times the median corner spacing, so
that they appear in the diagnostic views without entering the estimation.
The core can also be imported independently of the interface, which is how
the experiments of Section~\ref{sec:exp} invoke the same functions in
headless mode.

\end{document}